# Enhancing Affinity Propagation for Improved Public Sentiment Insights

Mayimunah Nagayi[1][0009-0009-9241-3059] and Clement Nyirenda[2][0000-0002-4181-0478]

Department of Computer Science, University of the Western Cape, Bellville, RSA

[1]4163113@myuwc.ac.za, [2]cnyirenda@uwc.ac.za

**Abstract.** With the large amount of data generated every day, public sentiment is a key factor for various fields, including marketing, politics, and social research. Understanding the public sentiment about different topics can provide valuable insights. However, most traditional approaches for sentiment analysis often depend on supervised learning, which requires a significant amount of labeled data. This makes it both expensive and time-consuming to implement. This project introduces an approach using unsupervised learning techniques, particularly Affinity Propagation (AP) clustering, to analyze sentiment. AP clustering groups text data based on natural patterns, without needing predefined cluster numbers. The paper compares AP with K-means clustering, using TF-IDF Vectorization for text representation and Principal Component Analysis (PCA) for dimensionality reduction. To enhance performance, AP is combined with Agglomerative Hierarchical Clustering. This hybrid method refines clusters further, capturing both global and local sentiment structures more effectively. The effectiveness of these methods is evaluated using the Silhouette Score, Calinski-Harabasz Score, and Davies-Bouldin Index. Results show that AP with Agglomerative Hierarchical Clustering significantly outperforms K-means. This research contributes to Natural Language Processing (NLP) by proposing a scalable and efficient unsupervised learning framework for sentiment analysis, highlighting the significant societal impact of advanced AI techniques in analyzing public sentiment without the need for extensive labeled data.

## 1 Introduction

Sentiment analysis, a subfield of natural language processing (NLP), also known as 'opinion mining', is a crucial area that focuses on extracting and analyzing sentiments, opinions, and emotions conveyed in textual data [1] [2]. Given the large increase in online content and social media platforms, analyzing sentiments has become fundamental for interpreting public opinion, customer feedback, and social trends [3]. The primary objective of sentiment analysis is to classify or predict words into predefined sentiment categories such as negative, neutral, or positive [2].

Usually, supervised learning approaches have been the predominant method for sentiment analysis. These methods involve training models on labeled datasets to predict and classify sentiment labels for unseen data [1]. In the past few years, there has been an increasing focus on investigating unsupervised learning methods for sentiment analysis [4]. Unsupervised learning approaches, including automated keyword extraction, do not rely on labeled data for training which makes them potentially more scalable and adaptable to different domains [2]. By uncovering hidden patterns and relationships, unsupervised learning algorithms can automatically classify text into sentiment categories [4].

This study is inspired by Zhang *et al.* [5], who examine the strengths and limitations of large language models (LLMs) in sentiment analysis, focusing on supervised learning. In contrast, this work proposes an unsupervised framework that utilizes Affinity Propagation (AP) and Agglomerative Hierarchical Clustering (AHC). This approach addresses Zhang's challenges, particularly the dependence on large labeled datasets, which are often costly and time-consuming to obtain. Iparraguirre-Villanueva *et al.* [4] note that while K-means clustering is commonly used in sentiment analysis, it requires predefined cluster numbers and may fail to capture complex sentiments in informal contexts like social media [3]. In comparison, AP dynamically determines the number of clusters based on data, offering a more nuanced view of public sentiment. Additionally, existing literature often neglects the integration of hierarchical clustering with unsupervised techniques [6]. By combining AP with AHC, this study enhances clustering quality and provides a hierarchical view of sentiment structures, marking a significant advancement in sentiment analysis by bridging traditional supervised methods with modern unsupervised techniques.

Reliance on supervised learning methods for sentiment analysis poses several challenges. While annotated datasets exist, obtaining them can be costly and time-consuming, particularly in specific domains and low-resource languages, limiting scalability across diverse areas [7]. Moreover, supervised models often struggle to generalize, leading to reduced accuracy with new textual data, which affects their applicability in dynamic fields like social research where data continuously emerge [8] [9]. Additionally, traditional approaches may fail to capture nuanced patterns in natural language, especially in informal contexts such as social media, where language is variable and context-dependent [10]. This results in misclassification and less insightful analysis. Therefore, there is an urgent need for scalable, cost-effective, and adaptable sentiment analysis methods that can operate without extensive labeled data and generalize across various domains and languages [7].

This research compares Affinity Propagation (AP) clustering and Agglomerative Hierarchical Clustering (AHC) with K-means clustering for text classification effectiveness [4]. It employs Term Frequency-Inverse Document Frequency (TF-IDF) Vectorization and Principal Component Analysis (PCA) for dimensionality reduction [11]. Integrating AP with AHC enhances sentiment analysis by providing hierarchical cluster representation [6]. However, AP's message-passing mechanism may struggle with nuanced sentiments in informal contexts like social media and could yield meaningless clusters without validation [12]. Metrics such as the Silhouette Score, Davies-Bouldin Index, and Calinski-Harabasz Score assess cluster quality [13]. This study contributes to sentiment analysis and unsupervised learning in NLP using datasets from the referenced study [5], emphasizing innovative AI applications.

The remainder of the paper is organized as follows: Section 2 reviews the algorithms used in this work, describing each algorithm. Section 3 outlines the proposed approach, detailing the methodology and implementation. Section 4 discusses the performance metrics used for evaluation. Section 5 discusses the results in detail, including the steps followed to obtain the results, ensuring fairness, and presenting the results and statistical analysis. Finally, Section 6 concludes the paper, summarizing the findings.

## 2      Unsupervised Learning Algorithms Used In This Work

This section provides a descriptive review of the algorithms utilized in this project, focusing on Affinity Propagation (AP) and Agglomerative Hierarchical Clustering(AHC ) for sentiment analysis. Additionally, K-Means clustering will be incorporated to assess the effectiveness of the proposed clustering methodologies.

### 2.1      Affinity Propagation (AP) Clustering

Affinity Propagation (AP) is an unsupervised clustering algorithm introduced by Frey and Dueck [14]. Unlike most traditional clustering methods that require the number of clusters to be specified before clustering, AP determines the number of clusters based on the data itself [15]. The algorithm operates by exchanging messages between data points until a set of exemplar points emerges. These exemplars are representative points that serve as the centers of clusters. AP is an exemplar-based clustering whereby it identifies representative points (exemplars) from a dataset and assigns other data points based on similarity, using an iterative message-passing process with responsibility and availability messages to update the likelihood of data points being exemplars or assigned to existing ones [14] [16].

AP has been utilized across various domains, including when Abdulah *et al.* [15] introduced an active clustering algorithm for data streams based on AP. Their work demonstrated the algorithm's effectiveness in managing evolving data streams by dynamically integrating new data points into existing clusters without the need to re-execute the entire clustering process. Recently, Castano *et al.* [16]  introduced an incremental extension of AP, known as A-Posteriori Affinity Propagation (APP). This method enhances scalability and adaptability by integrating new data points into existing clusters and allowing for the removal of outdated clusters. Additionally, Guan *et al* [12] applied AP for clustering text in seed construction.

Fig. 1 illustrates the iterative process of the Affinity Propagation (AP) clustering algorithm. Here's a breakdown of what each part of the image shows [14]:

- **Initialization**: The process begins with all data points, marked in green, without any exemplars identified.

- **Iterations 1 to 6**: As the iterations progress, the algorithm communicates information through messages between data points. These messages update the likelihood of each point being an exemplar or being assigned to an existing one. The connections and colors start to change, indicating the evolving relationships and the emerging exemplars.
- **Convergence**: By the final stage, exemplars are identified and marked in red. The data points are grouped around these exemplars, forming distinct clusters.

The arrows between iterations represent the passage of time and the iterative nature of the algorithm, showing how the clustering structure evolves until convergence [14].

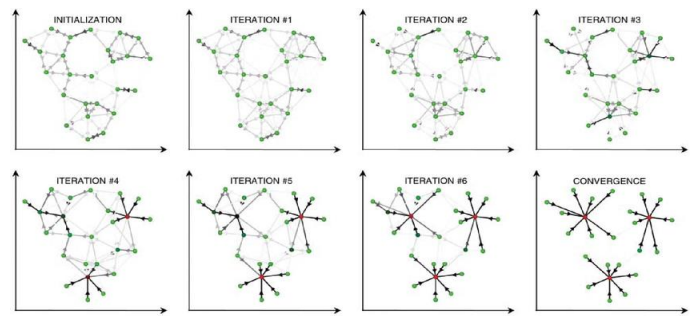

**Fig. 1.** The iterative process of Affinity Propagation shows the emergence of exemplars (red) and clusters through message passing over several iterations [14].

### 2.2 Agglomerative Hierarchical Clustering

Agglomerative Hierarchical Clustering (AHC ) is a hierarchical clustering method that builds clusters by continuously combining the most similar data points [17]. It starts with each data point as an individual cluster and merges them based on similarity measures until all points belong to a single cluster or a predefined number of clusters is reached. This method is particularly useful for creating a hierarchical structure of the data, which can be visualized using a 'dendrogram' [17] [18].

AHC has been utilized in sentiment analysis alongside other methods, such as lexicon-based approaches, to evaluate the sentiments of Twitter users regarding the revision of Indonesian laws [19]. AHC has also been applied to classify large datasets derived from field data related to pop-up housing environments which helped in understanding and optimizing the use of temporary housing solutions in urban areas [6].

### 2.3 K-Means Clustering

K-Means Clustering is a commonly used clustering algorithm that partitions data into K clusters based on similarities. It is particularly popular due to its efficiency, simplicity, and ease of implementation. The algorithm iteratively assigns data points to clusters and updates cluster centers until convergence, making it suitable for various applications, including sentiment analysis [4] [8].

The K-means algorithm has been applied to various topics. For example, Iparraguirre-Villanueva *et al.* [4] conducted a comprehensive study on sentiment analysis of tweets using K-means and unsupervised learning techniques. Their research delved into the classification and analysis of Twitter (Known as X) content related to the Pension and Funds Administration (AFP) [4].

## 3 Proposed Approach

This section details the approach to discovering public sentiment using Affinity Propagation (AP) clustering. The method leverages the strengths of AP in identifying sentiment patterns from tweet data. The proposed approach consists of several key steps including data preprocessing, feature extraction, clustering using AP with AHC, and evaluation of the clustering results as shown in Fig. 2.

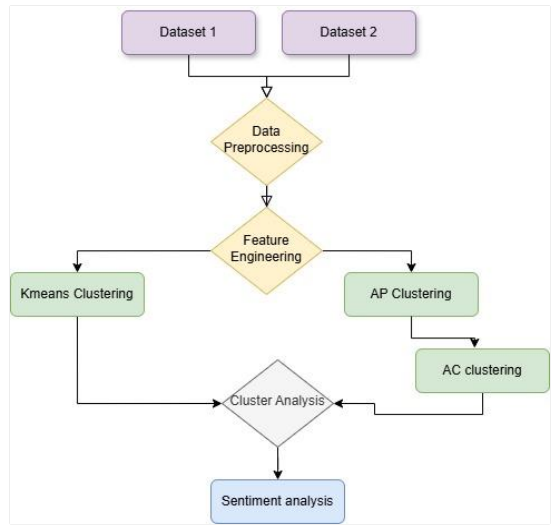

**Fig. 2.** Proposed Algorithmic Framework for Sentiment Analysis Using Affinity Propagation and Agglomerative Clustering.

### 3.1 Data Preprocessing

This study uses two main datasets:

*a) Twitter Data from Zhang et al.* [5]*:* This dataset comprises 500 unique tweets collected from various topics relevant to public sentiment. Each entry in this dataset includes columns such as *original_index, sid, text*, and *label*. The text column contains the tweet content, while the label indicates the sentiment classification (e.g., positive, negative, neutral). This dataset serves as a foundation for understanding how public sentiments are expressed in concise formats typical of social media interactions.

*b) Kaggle Twitter Dataset* [20]*:* This dataset consists of approximately 27,500 tweets labeled according to their sentiment (positive, negative, neutral). It includes columns such *as textID, text, selected_text*, and *sentiment*. The selected_text column highlights specific phrases within the tweet that encapsulate the overall sentiment. This dataset is valuable for developing an NLP-based classifier to identify negative tweets and reduce harmful content on social media. Tweets from diverse events and discussions on Twitter provide rich sentiment analysis. The preprocessing step includes merging datasets, removing noise (e.g., URLs, hashtags), handling missing values, and normalizing text through tokenization, lowercasing, and removing stop words.

### 3.2 Feature Extraction

Term Frequency-Inverse Document Frequency (TF-IDF) vectorization is used to transform the textual data into a numerical format suitable for clustering. TF-IDF captures the importance of terms within the tweets, balancing term frequency with inverse document frequency to highlight significant words while minimizing the impact of common terms [21]. While TF-IDF may oversimplify text representation, it was chosen for its effectiveness in highlighting significant terms within the dataset, which is crucial for comparing clustering methods in this study.

### 3.3 Clustering With K-Means(Baseline)

The K-means clustering algorithm is implemented using the '*KMeans*' function from the Python Scikit-learn library. This function is initialized with the preprocessed TF-IDF numerical vector data from the Tweets, which was prepared in the data preparation Jupyter notebook. It is configured with three clusters for subsequent sentiment classification. The K-means algorithm follows these steps [4]:
- *Centroid Initialization:* K-means start by randomly initializing centroids for each cluster.
- *Assignment Step:* Each data point is assigned to the nearest centroid, forming the algorithm's initial clusters.
- *Centroid Update:* The algorithm recalculates the centroids by averaging the data points assigned to each cluster.
- *Iteration:* The assignment and update steps are repeated until the centroids no longer change, indicating convergence.

After clustering, the labels assigned by K-means are stored for further analysis in CSV and Pickled Python Objects(pkl) formats.

### 3.4 Clustering With Affinity Propagation

The Affinity Propagation algorithm is implemented using the built-in *'AffinityPropagation'* function from the Scikit-learn library in Python. This function is initialized with the preprocessed Tweets numerical vector data and is configured with parameters optimized for the dataset, such as:
- Maximum Number of Iterations: Set to **500** to ensure that the algorithm has sufficient opportunities to converge. This parameter was chosen based on preliminary tests that indicated it allows for adequate refinement of clusters without excessive computational overhead.
- Damping Factor: Set to **0.9** to control the convergence process. The damping factor helps stabilize the updating process by reducing oscillations that could prevent convergence. This value was determined through iterative testing, where different damping factors were evaluated for their impact on clustering stability and quality.

These parameters were selected to balance computational efficiency and clustering quality, ensuring that the algorithm produces reliable results across multiple runs. The effectiveness of these settings was validated by observing clustering outcomes during initial trials, which aimed to achieve consistent and accurate clustering results.

The algorithm iteratively updates responsibility and availability messages between tweets until convergence [16]. The built-in Affinity Propagation function provides an efficient and optimized implementation of the algorithm, handling tasks such as:
- *Similarity Calculation:* Computing the similarity between all pairs of tweets using the negative squared Euclidean distance [22].
- *Responsibility Update:* Reflecting how well-suited a tweet is to be an exemplar for another tweet, considering other potential exemplars [14] [22].
- *Availability Update:* Assessing how well a tweet chooses an exemplar based on the preferences expressed by other tweets [12] [22].
- *Cluster Assignment:* Identifying exemplars and forming clusters around these exemplars based on the highest combined responsibility and availability scores [15] [22].
- *Convergence Checks:* Iteratively updating responsibility and availability messages until convergence [22].

The labels assigned to each data point by AP are collected and stored for further analysis in a CSV and pkl formats. After all data has been processed and their labels recorded, the AP data is then kept to further refine the clustered results by applying Agglomerative hierarchical Clustering.

Principal Component Analysis (PCA) with 100 components was employed to visualize the clustering results.

### 3.5 Integration With Agglomerative Hierarchical Clustering

To improve the interpretability and clustering quality of sentiment analysis, this work combines Agglomerative Hierarchical Clustering (AHC) with Affinity Propagation (AP) clustering. While AP dynamically identifies exemplars within the data, AHC offers a hierarchical view by iteratively merging the most similar clusters [23].

a. *Initial Clustering with AP:* The algorithm first identifies exemplars from the preprocessed tweet data. These exemplars represent initial clusters based on the highest combined responsibility and availability score [22].

b. *Hierarchical Refinement*: The exemplars derived from AP are then subjected to AHC. This method merges the most similar clusters iteratively until a predefined number of clusters is achieved. In this project, a predefined cluster count of three is used to align with the cluster count in K-means, ensuring a fair comparison.

c. *Similarity Measurement:* Agglomerative Hierarchical Clustering utilizes the same TF-IDF vectorized data and negative squared Euclidean distance for calculating similarities between clusters. This consistency in similarity measurement ensures that the hierarchical clustering builds upon the foundational structure identified by AP.

## 4  The Performance Metrics

The effectiveness of the clustering algorithms was evaluated using the Silhouette Score, Calinski-Harabasz Score, and Davies-Bouldin Index. A detailed explanation of each metric, including its purpose and implementation using built-in functions, is provided next.

### 4.1  Silhouette Score

The Silhouette Score measures how similar an object is to its own cluster compared to other clusters. It ranges from -1 to 1 as [-1; 1], with higher values indicating better-defined clusters. A score close to 1 suggests that the data points are well-matched to their cluster and poorly matched to neighboring clusters, while a score near -1 indicates that data points might have been assigned to the wrong cluster [13] [24]. A higher Silhouette Score indicates distinct sentiment patterns, making it particularly useful for applications in marketing where understanding consumer sentiment can drive targeted campaigns. The Silhouette Score was calculated using the built-in 'silhouette_score' function from the Scikit-learn library. To manage computational resources, the score for both K-means and the combination of Affinity Propagation (AP) with Agglomerative Hierarchical Clustering (AHC ) was computed on a subset of the data. The TF-IDF matrix and the corresponding cluster labels served as inputs for these calculations.

### 4.2  Calinski-Harabasz Score

The Calinski-Harabasz Score, also known as the Variance Ratio Criterion, evaluates the ratio of the sum of between-cluster dispersion to within-cluster dispersion. Higher scores indicate better-defined clusters with a range of [0; +]. This metric measures the compactness and separation of the clusters, with higher values signifying that clusters are dense and well-separated [13]. This is essential for social research applications where nuanced understanding of public opinion is required. The Calinski-Harabasz Score was computed using the built-in *'calinski_harabasz_score'* function from the

Scikit-learn library. For both K-means and the combination of AP with AHC, the score was calculated using the TF-IDF matrix and the cluster labels. A subset of the data was used to ensure efficient computation.

### 4.3 Davies-Bouldin Index

The Davies-Bouldin Index measures the average similarity ratio of each cluster with its most similar cluster, where lower values indicate better clustering. This index is calculated as the average ratio of intra-cluster distance to inter-cluster distance, as a range [0; +] with lower values indicating that clusters are compact and well-separated from each other [13]. Lower Davies-Bouldin Index values highlight effective segmentation of customer feedback, crucial for organizations aiming to distinguish between different sentiment groups effectively. The Davies-Bouldin Index was calculated using the built-in 'davies_bouldin_score' function from the Scikit-learn library. For both K-means and the combination of AP with AHC , the score was computed using the TF-IDF matrix and the cluster labels. A subset of the data was used to facilitate the calculation.

### 4.4 Execution Time

Execution time is a key performance metric, particularly with large datasets. Comparing the elapsed time for K-means and the combined AP and AHC algorithms offers valuable insights into their computational efficiency. Python's built-in time functions were used to record the execution time of both methods, helping to assess their computational cost and scalability.

## 5 Results And Discussion

The following results were obtained after analyzing the performance of the clustering algorithms. The dataset comprised cleaned tweets, vectorized using TF-IDF, resulting in a reduced feature matrix of shape (27,981, 100) from a full feature matrix of shape (27981, 28645). The computations were performed on a machine with the following specifications: 64-bit OS with an x64-based processor, 8.00GB installed RAM, and an Intel(R) Core i5-6400 CPU running at 2.70GHz. Each clustering algorithm was executed more than 10 times (recorded 13 times) to ensure fairness and reliability of the results. The K-means results are visualized using Principal Component Analysis (PCA) with 100 components. Fig. 3 illustrates the clusters identified by the K-means algorithm, where different colors represent the 3 distinct clusters, demonstrating how K-means has grouped similar data points based on their similarity. Figure 4 provides insights into the distribution and separation of the clusters formed by the algorithm. The resulting clusters from the Affinity Propagation algorithm with agglomerative hierarchical clustering are presented in Figure 4. Each point represents a data sample, and the colors indicate the different clusters.

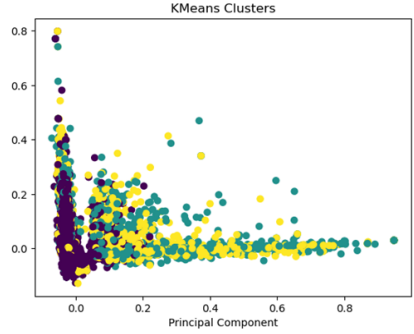
**Fig. 3.** K-means clusters using PCA.

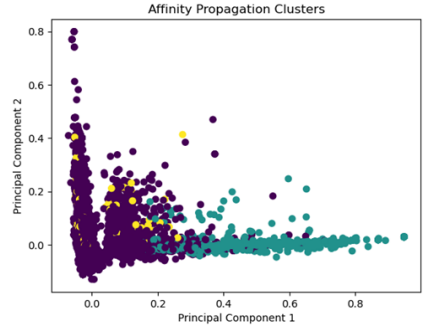
**Fig. 4.** Affinity Propagation clusters using PCA.

Table 1 presents the execution time, Silhouette Scores, Calinski-Harabasz Scores, Davies-Bouldin Index, and the execution time for both the K-means algorithm, Affinity Propagation with Agglomerative Hierarchical Clustering (AP & AHC), and Affinity Propagation alone (AP Only).

**Table 1.** Comparative Evaluation of the Algorithms.

| Metric | K-means | AP & AHC | AP(Only) |
|---|---|---|---|
| Silhouette Score | -0.333 | **0.173** | 0.113 |
| Calinski-Harabasz Score | 0.971 | **14.596** | 9.015 |
| Davies-Bouldin Index | 5.334 | 1.961 | **0.985** |
| Time (seconds) | 301.50 | 456.75 | **49.763** |

The analysis shows that K-means demonstrates superior speed due to its single algorithmic approach. However, Affinity Propagation with Agglomerative Hierarchical Clustering, despite its complexity, results in better clustering quality metrics. While K-means is more efficient in execution time, averaging 301.50 seconds, Affinity Propagation with Agglomerative Hierarchical Clustering takes longer, averaging 456.75 seconds. Notably, when considering Affinity Propagation alone, which averages 49.763 seconds, it is clear that the longer execution time is primarily due to the added processing of Agglomerative Hierarchical Clustering.

Although Affinity Propagation offers a relatively quick execution time, it tends to produce many clusters because it identifies exemplars based on data characteristics. This can lead to an excessive number of clusters that may not correspond to meaningful sentiment categories, making it less ideal for sentiment analysis applications where interpretability is crucial. While Affinity Propagation's standalone efficiency highlights its effectiveness within a reasonable timeframe, its propensity to generate numerous clusters requires careful consideration when specific sentiment categories are desired.

## 5.1 Statistical Analysis

The mean of the metrics was calculated for each algorithm to summarize the central tendency and variability of the results. A two sample t-test was conducted to compare the performance of K-means and Affinity Propagation with Agglomerative Hierarchical Clustering. A two-sample t-test was employed to evaluate the Silhouette Score, Calinski-Harabasz Score, and Davies-Bouldin Index in order to assess whether there are significant differences between the two clustering methods.

- **Null Hypothesis ($H_0$):** There is no significant difference in the clustering quality metrics between K-means and the combination of Affinity Propagation with Agglomerative Hierarchical Clustering.
- **Alternative Hypothesis ($H_1$):** There is a significant difference in the clustering quality metrics between K-means and the combination of Affinity Propagation with Agglomerative Hierarchical Clustering.

The results of the two sample t-test are summarized in Table 2.

**Table 2.** Statistical analysis summary table

| Metric | t-statistic | p-value |
| --- | --- | --- |
| Silhouette Score | -5.895 | $4.416 \times 10^{-0.6}$ |
| Calinski-Harabasz Score | -84.375 | $3.341 \times 10^{-31}$ |
| Davies-Bouldin Index | 18.634 | $8.659 \times 10^{-16}$ |

Statistical analysis using the paired t-test reveals significant differences ($p < 0.001$) between K-means and the combination of Affinity Propagation with Agglomerative Hierarchical Clustering across all metrics. This clearly shows that the proposed approach outperforms the K-means approach. The discussion of the results for each of these metrics is presented as follows:

- **Silhouette Score.** The Affinity Propagation with Agglomerative Hierarchical Clustering achieves a higher silhouette score (0.17) compared to K-means (-0.33), indicating better-defined clusters. The negative silhouette score for K-means suggests poor clustering.
- **Calinski-Harabasz Score.** Affinity Propagation with Agglomerative Hierarchical Clustering has a significantly higher Calinski-Harabasz Score (14.60) than K-means (0.97), suggesting a better-defined cluster structure with greater separation between clusters and tighter cluster formation.
- **Davies-Bouldin Index(DBI).** The lower Davies-Bouldin Index (1.96) for Affinity Propagation with Agglomerative Hierarchical Clustering compared to K-means (5.33) indicates more distinct clusters. Lower DBI values suggest that clusters are well-separated and compact.

The performance of Affinity Propagation with Agglomerative Hierarchical Clustering over K-means can be attributed to the underlying differences in how these two algorithms operate. When combined with Agglomerative Hierarchical Clustering, Affinity Propagation is more effective in capturing the complex patterns in sentiment data.

It identifies natural groupings without requiring a pre-specified number of clusters, allowing it to adapt better to the data's inherent structure.

On the other hand, K-means assumes that clusters are evenly distributed and requires the number of clusters to be set in advance. This assumption can lead to poorer performance, especially in sentiment analysis, where the data often displays more difficult relationships. The negative silhouette score for K-means highlights its struggle to properly classify the data, leading to poorly defined clusters.

In contrast, the higher Calinski-Harabasz Score and lower Davies-Bouldin Index observed with Affinity Propagation and Agglomerative Hierarchical Clustering suggest that this combination forms more distinct, well-separated clusters, reflecting its ability to better capture the underlying sentiment patterns. These differences emphasize the advantage of using methods that can dynamically adapt to the data's natural structure, particularly in tasks like sentiment analysis.

### 5.2 Clustering Results

The experimental results summarized in Table 1 show that AP with AHC outperforms K-means in clustering quality, as indicated by higher Silhouette Scores, a higher Calinski-Harabasz Score, and lower Davies-Bouldin Indices. The last two tweets from each cluster are provided to examine the clustering algorithms' results. Table 3 and Table 4 present insights into the cluster assignments generated by K-means and AP & AHC.

**Table 3.** The last 2 cluster Samples from K-means Clusters.

| Cluster | Text/Tweet | Sentiment score |
|---|---|---|
| 1 | really im practicing vegetarianism ive seen little results | {'neg': 0.0, **'neu': 1.0,** 'pos': 0.0, 'compound': 0.0} |
| 1 | trash put kids head shame youor idiot probably seems run family | {'neg': 0.0, 'neu': 0.303, **'pos': 0.697,** 'compound': 0.9442} |
| 2 | great visit st brigids callan today met great students good rivalry tipp kk girls | {'neg': 0.0, **'neu': 0.518,** 'pos': 0.482, 'compound': 0.8519} |
| 2 | wait burger king lax good idea hour flight didnt anyone stop nationalfastfoodday | {'neg': 0.303, **'neu': 0.526,** 'pos': 0.171, 'compound': -0.25} |
| 3 | go bed muhammad ali | {'neg': 0.0, **'neu': 1.0,** 'pos': 0.0, 'compound': 0.0} |
| 3 | uh oh bad hombres makes ya go hhmmm | {**'neg': 0.61**, 'neu': 0.39, 'pos': 0.0, 'compound': -0.5719} |

**Table 4.** The last 2 cluster Samples from AP with AHC Clusters.

| Cluster | Text/Tweet | Sentiment score |
|---|---|---|
| 1 | id responded going | {'neg': 0.0, **'neu': 1.0,** 'pos': 0.0, 'compound': 0.0} |
| 1 | failed inspection know pass wooven woantitip bracket sold woven worse taxes | {'neg': 0.416, **'neu': 0.584,** 'pos': 0.0, 'compound': -0.7506} |
| 2 | happy mothers day mums | {'neg': 0.0, 'neu': 0.448, **'pos': 0.552,** 'compound': 0.5719} |
| 2 | happy mothers day mothers world | {'neg': 0.0, **'neu': 0.519,** 'pos': 0.481, 'compound': 0.5719} |
| 3 | watching thankyouobama u make wan na remix using sweet sable old times sake beat unexpected | {'neg': 0.0, **'neu': 0.812,** 'pos': 0.188, 'compound': 0.4588} |
| 3 | wan na sue electoral college someone isi asked could class action u r legal eaglecan | {'neg': 0.0, **'neu': 0.889,** 'pos': 0.111, 'compound': 0.128} |

### 5.3 Sentiment Analysis

The analysis of sentiment within the clusters reveals distinct emotional tones captured by both clustering methods, K-means and Affinity Propagation with Agglomerative Hierarchical Clustering (AP & AHC). The classification of clusters based on sentiment is as follows:

a. **K-means Clusters:**
- **Cluster 3 (Neutral):** This cluster primarily consists of neutral statements that reflect logistical updates rather than emotional content. For instance, tweets like "really I'm practicing vegetarianism I've seen little results" indicate a focus on personal choices without conveying strong feelings.
- **Cluster 2 (Positive):** This cluster includes positive sentiments related to experiences and interactions. Tweets such as "great visit st brigids callan today met great students, good rivalry tipp kk girls" express enjoyment and appreciation, reflecting a favorable emotional tone.
- **Cluster 1 (Negative):** This cluster exhibits strong negative sentiments, indicating dissatisfaction or concern. An example is the tweet "uh oh bad hombres makes ya go hhmmm," which suggests unease about a situation.

b. **AP with AHC Clusters:**
- **Cluster 3 (Neutral):** This cluster primarily consists of neutral statements that reflect logistical updates rather than emotional content. For instance, tweets like "really I'm practicing vegetarianism I've seen little results" indicate a focus on personal choices without conveying strong feelings.
- **Cluster 2 (Positive):** This cluster includes positive sentiments related to experiences and interactions. Tweets such as "great visit st brigids callan today met great students, good rivalry tipp kk girls" express enjoyment and appreciation, reflecting a favorable emotional tone.
- **Cluster 1 (Negative):** This cluster exhibits strong negative sentiments, indicating dissatisfaction or concern. An example is the tweet "uh oh bad hombres makes ya go hhmmm," which suggests unease about a situation.

### 5.4 Sentiment Scores

Sentiment scores quantify the emotional tone of text data, classifying sentiments as positive, negative, or neutral based on specific words or phrases. For example, "good" and "happy" yield positive scores, while "bad" and "sad" result in negative scoresscores [1]. Limitations of Sentiment Scores includes:
- *Word-Based Analysis:* Many algorithms rely on keyword recognition, meaning the presence of certain words can skew sentiment scores without considering the context. For instance, "bad" might not indicate negativity in a sarcastic remark.
- *Lack of Contextual Understanding:* Sentiment analysis tools often struggle with nuances like irony or cultural references, leading to misclassification based on superficial word matching.
- *Simplification of Complex Emotions:* Human emotions are intricate; sentiment scores can oversimplify them into basic categories. For example, a tweet expressing frustration may receive a negative score but could also imply humor or resignation that isn't captured.

The comparison of samples from both clustering methods highlights the effectiveness of AP combined with Agglomerative Hierarchical Clustering (AHC) in capturing intricate sentiment structures within social media texts. While K-means provides basic clustering based on established patterns, AP & AHC enhance sentiment analysis by recognizing both emotional highs and lows, revealing the underlying complexities of public sentiment. K-means clusters are categorized as neutral, positive, and negative, while AP & AHC clusters demonstrate a more nuanced approach with negative, positive, and neutral classifications. This improved capability aligns with the project's objective of providing deeper insights without relying heavily on labeled data, showcasing the value of unsupervised learning techniques in sentiment analysis. By effectively identifying and preserving the emotional nuances present in text data, AP & AHC not only improves clustering quality but also offers a hierarchical view of sentiment structures, making it a significant advancement in understanding public sentiment.

## 6 Conclusions

This project explores the potential of unsupervised learning in sentiment analysis of tweets, with a focus on Affinity Propagation (AP) combined with Agglomerative Hierarchical Clustering. The objective was to compare unsupervised methods in sentiment analysis tasks, contributing to advancements in social media analytics and customer sentiment analysis methodologies. By combining two datasets, the paper enhances the model's learning experience and provides a framework for analyzing public sentiment without the need for extensive labeled data. The primary goal was to compare AP with traditional K-means clustering and the results demonstrate that AP clustering, especially when combined with Agglomerative Hierarchical Clustering, significantly outperforms K-means in forming coherent and distinct sentiment clusters. The effectiveness of these methods was evaluated using the Silhouette Score, Calinski-Harabasz Score, and Davies-Bouldin Index. AP with Agglomerative Hierarchical Clustering achieved a Silhouette Score of 0.173, a Calinski-Harabasz Score of 14.596, and a Davies-Bouldin Index of 1.961, compared to K-means scores of -0.333, 0.971, and 5.334, respectively. These results highlight the hybrid approach's superior performance in effectively capturing global and local sentiment forms. This research contributes to the field of sentiment analysis by proposing a scalable and efficient unsupervised learning framework that enhances our understanding of public sentiment without the extensive requirement for labeled data.

Future work could explore several avenues to enhance the effectiveness of the proposed unsupervised learning approach. Expanding data sources beyond Twitter to include reviews from e-commerce platforms, forums, or news articles would improve robustness and provide a more comprehensive view of public sentiment. Additionally, incorporating contextual information such as user demographics and temporal factors is essential for achieving better clustering results. Evaluating these proposed unsupervised techniques against state-of-the-art supervised models would yield valuable insights into their strengths and weaknesses in real-world applications. Implementing these methods in real-world scenarios will also validate their practicality in monitoring public sentiment during significant events or crises. Lastly, refining clustering algorithms by exploring advanced techniques, including deep learning-based approaches or hybrid models that integrate supervised learning elements, could further enhance sentiment classification accuracy.